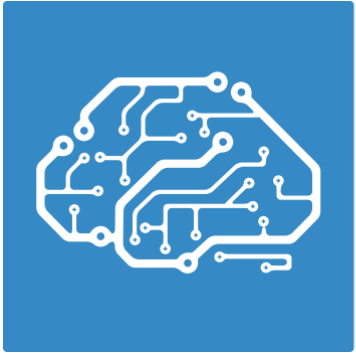

# When Computer Vision Gazes at Cognition

by
Tao Gao*, Daniel Harari*, Joshua Tenenbaum and Shimon Ullman

**Abstract:** Joint attention is a core, early-developing form of social interaction. It is based on our ability to discriminate the third party objects that other people are looking at. While it has been shown that people can accurately determine whether another person is looking directly at them versus away, little is known about human ability to discriminate a third person gaze directed towards objects that are further away, especially in unconstraint cases where the looker can move her head and eyes freely. In this paper we address this question by jointly exploring human psychophysics and a cognitively motivated computer vision model, which can detect the 3D direction of gaze from 2D face images. The synthesis of behavioral study and computer vision yields several interesting discoveries. (1) Human accuracy of discriminating targets 8°-10° of visual angle apart is around 40% in a free looking gaze task; (2) The ability to interpret gaze of different lookers vary dramatically; (3) This variance can be captured by the computational model; (4) Human outperforms the current model significantly. These results collectively show that the acuity of human joint attention is indeed highly impressive, given the computational challenge of the natural looking task. Moreover, the gap between human and model performance, as well as the variability of gaze interpretation across different lookers, require further understanding of the underlying mechanisms utilized by humans for this challenging task.

* Tao Gao and Daniel Harari contributed equally to this work.

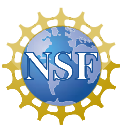

This work was supported by the Center for Brains, Minds and Machines (CBMM), funded by NSF STC award CCF-1231216.

# When Computer Vision Gazes at Cognition

Tao Gao\*, Daniel Harari\* †,
Joshua Tenenbaum\*, Shimon Ullman\* †

\* Center for Brains, Minds and Machines, Massachusetts Institute of
Technology, Cambridge, MA
† Weizmann Institute of Science, Rehovot, Israel

## Abstract

Joint attention is a core, early-developing form of social interaction. It is based on our ability to discriminate the third party objects that other people are looking at. While it has been shown that people can accurately determine whether another person is looking directly at them versus away, little is known about human ability to discriminate a third person gaze directed towards objects that are further away, especially in unconstraint cases where the looker can move her head and eyes freely. In this paper we address this question by jointly exploring human psychophysics and a cognitively motivated computer vision model, which can detect the 3D direction of gaze from 2D face images. The synthesis of behavioral study and computer vision yields several interesting discoveries. (1) Human accuracy of discriminating targets 8°-10° of visual angle apart is around 40% in a free looking gaze task; (2) The ability to interpret gaze of different lookers vary dramatically; (3) This variance can be captured by the computational model; (4) Human outperforms the current model significantly. These results collectively show that the acuity of human joint attention is indeed highly impressive, given the computational challenge of the natural looking task. Moreover, the gap between human and model performance, as well as the variability of gaze interpretation across different lookers, require further understanding of the underlying mechanisms utilized by humans for this challenging task.

## 1    Introduction

### 1.1    Understanding social agency in real scenes

As a social species, humans are remarkably good at understanding other's mental states based on visual perception alone. Recent work on social perception have been done in 2D toy world with cartoon like characters (e.g., [1], [2]). In the meanwhile, there has been a growing interest in understanding non-verbal social interactions in real scenes. Many developmental studies have demonstrated that even young infant can understand others' mental states by observing their non-verbal actions [3]–[5]. Nevertheless, this type of visual social understanding poses a challenge for both cognitive sciences and computer vision. The reason is that most human social interactions occur in real scenes, in which agents act freely. However, due to the complex interactions between the 3D environment and human actions, analyzing these scenes can be extremely challenging.

Addressing the above challenge encourages an inter-disciplinary research between cognitive science and computer vision. From the cognitive science perspective, the advances of state-of-the-art computer vision techniques, facilitate rigorous studies on human's spontaneous social behaviors. More importantly, "grounding" a social process onto real images can force researchers to consider computational challenges that may be absent in cartoon worlds. From the computer vision perspective, working on cognitively motivated questions can lead the field move toward human-like rich scene understanding, beyond object recognition (e.g. building a vision model that can track the 'false-belief' of an observed person, given what this person can or cannot see in the visual scene [6], [7]).

## 1.2   The perception of "free" gaze

To achieve the long-term goal of understanding social agency in real scenes, one initial step is to interpret a person's gaze direction, which is a window to one's mental states. This ability is the foundation of human's joint attention, and many other important social interactions. Infants begin to develop this ability [8], which plays an important role in development of communication and language [3], as early as at the age of 3-6 months [5].

Gaze perception has been extensively studied in human perception since 1960s [9]–[15]. However, there are several interesting contrasts between the scope of existing behavioral studies and the understanding of gaze perception in real scenes. First, most behavioral studies have been focusing on judging whether the gaze is directly towards the observers or not [9], [12]. The acuity of detecting direct eye-contact can be as high as 3° of visual angle [9], [10]. In a few studies in which the gaze is away from the observers' head, all possible directions are still surrounding a close vicinity and centered around the observer (e.g., in [10], a circle of 12° of visual angle around the observer), which only covers a very small the space in a scene. Secondly, almost all studies imposed strong constraints on the looker's gaze behavior, such as fixing the head at a certain pose and only allowing the rotation of the eyes. While this certainly helps to isolate the effects of head and eye orientations, it is unclear how to apply this results in natural looking scenes, in which the looker moves her head and gaze freely [16]. Thirdly, as developmental studies have shown, gazing towards an object is an important source for infant's learning. However, in most research on discriminating gaze perception, the threshold is measured by asking the looker to look at an empty space [9], [10]. This is again, different than natural and realistic situations, in which the person's gaze is typically oriented at an object in the scene.

In the field of computer vision, head pose and eye gaze estimations have been studied for a long time [17]–[19]. The vast majority of these studies addressed either the head pose estimation problem or the eye gaze tracking as disjoint problems. While these two research areas demonstrate an impressive performance for each of the separate tasks, very little was done addressing the problem of detecting the direction of gaze in natural and unconstrained scenes, where observed humans can look freely at different targets. In this natural looking settings, gaze events towards different targets may share the same head or eye poses, while different poses may share the gaze towards the same target. A recent work [20] suggested to estimate gaze under free-head movements by combining depth and visual data from a Microsoft Kinect sensor. In order to reduce the complexity of the gaze problem, eye images were rectified as if seen from a frontal view, by registering a person-specific 3D face model and tracking the head under different poses. However, the eyes rectification procedure is less likely to be cognitively plausible and sets limitation on the range of supported head poses (due to self-occlusions at non-frontal poses). This approach is also not suitable to predict 3D direction of gaze given only 2D images, as it requires the 3D head pose information during testing.

## 1.3   Contrasting Human and Computer Vision Performance

While having a threshold number on the acuity of human's gaze following skill is certainly important, we also hope to evaluate human performance by contrasting it with state-of-the-art computer vision models. This is a critical component of this study, as we are eager to know how challenging the perception of natural gaze actually is. Researchers usually don't have a good intuition about the difficulty of a cognitive process until they starts to reverse-engineer it. For instance, people on the street may feel that solving a differential equation is remarkably more intelligent than acquiring a common-sense knowledge of a 3-years old girl. It is actually the latter that is much more challenging for a machines to learn [21]. From the perspective of cognitive studies, we would like to focus more on aspects of social perception in which machine still cannot reach human-level performance. In addition, previous studies employed highly-constrained settings to avoid the variance in the experiment. Exploring the social perception in natural scenes will inevitably introduce a larger variance into the results. We hope the computational models can provide explanations to the variance. From the perspective of machine intelligence and computer vision, we would like to model the underlying mechanisms of gaze following in order to reach human performance on this task. This will provide AI systems with a powerful human-like ability for both scene and social understanding.

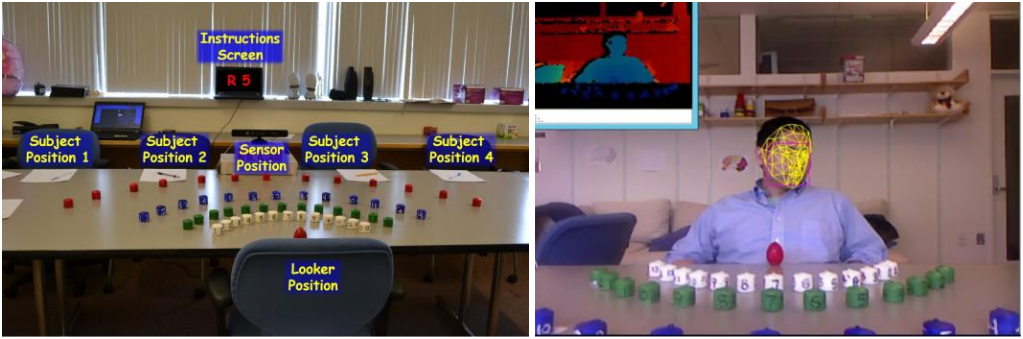

Figure 1: Experimental settings. (Left) The layout of the experiments consists of 52 objects arranged on a table in 4 concentric arcs, relative to the looker position. A Microsoft Kinect RGB-D sensor and a command screen are aligned in-front of the looker. 4 human subjects are positioned on the other side of the table, facing the looker, and noting their best guesses of the looker's gaze target. (Right) The Kinect sensor provides a RGB color image and a depth image. The Kinect's face tracking algorithm provides additional information of the 3D head pose orientation and facial feature locations.

The rest of this paper is laid-out as follows: In Section 2, we will describe our free-looking task for both human observers and computational models. In Section 3, we will introduce our appearance-based model which is able to predict a 3D gaze direction from a 2D image. In Section 4, we will compare human and model performance side by side, which provides valuable data for both cognitive sciences and computer vision. In Section 5, we will discuss the implications of our results to human and machine gaze perception in general.

## 2    Experimental Settings

In a free-looking task, a looker sits in a chair, facing an array of 52 objects. On the other side of the table (121cm), four human observers sit on chairs, facing the looker. In each trial, the looker gazes at an object by following a command presented on the monitor of a laptop located behind the observers. The action scene is recorded by a Microsoft Kinect Sensor with RGB-D cameras (Figure 1). In this setup, the looker, the center of the object array, the Kinect sensor and the screen of the command computer are all aligned in order, so that the looker will face directly at the camera between every two trials. The observers' task is to follow the looker's gaze and to write down their best guesses of the looker's target. The object array is laid as a concentric configuration with 13 columns consisting of 4 objects (width: 4.6cm; height: 3.6cm). The color (white, green, blue or red) and the number (1 through 13) painted on each object represent the object's row and column respectively. The radius of each row is 29.4, 49.7, 60.6 and 96.1cm respectively. In this configuration the visual angles between every two rows is 10° given the center of the two eyes is 35cm above the table. The angular difference between each two columns is 10° on the table surface. The corresponding visual angle is about 8°. In practice, the visual angles varies given the position of the head and can be computed on a trial-by-trial basis. The left most observer position (Position 1) is 180cm away from the looker. The angle between the looker's resting face direction and the direction to this position is 47.7° (angular distance). Position 2 is 138cm away from the looker, with an angular distance of 28.6°. Position 4 and Position 3 are mirror reflections of Position 1 and Position 2 over the center of the table respectively.

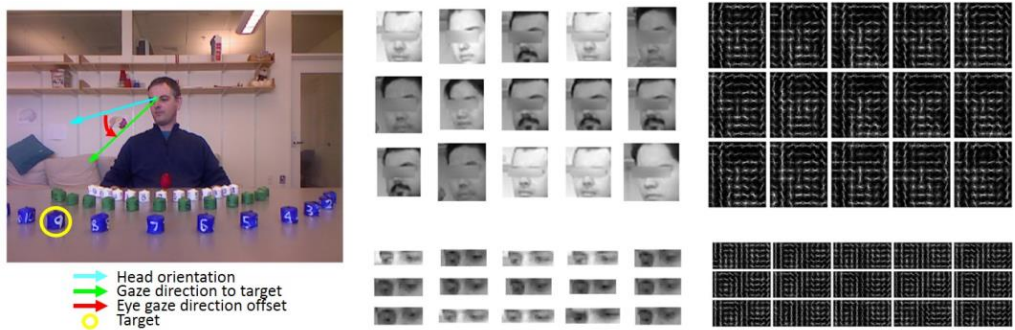

Figure 2: Computational appearance-based model. (Left) In each of the gaze trials the Appear-Face-Eyes model associates the face appearance with the head pose orientation (cyan), and the eyes appearance with the eye-gaze correction offset (red). The ground truth gaze direction (green) is the direction from the center of the eyes towards the target (yellow). (Middle) Extracted nearest neighbors for both face and eyes during a query event. Each neighbor is associated with an orientation, and the final direction is determined as the weighted average of these orientations, where the weights correspond to the neighbors' similarity with the query appearance. (Left) Rendered images of the Histogram of Gradients (HoG) descriptors used to represent face and eyes appearance. These robust general purpose image descriptors capture both edge and texture properties and allow spatial tolerance.

In Experiment 1, two Asian lookers (TG and JP) and 16 Asian observers participated. All with normal or corrected-to-normal vision. The observers were evenly assigned into 4 sections, each of which containing 8 blocks. To estimate the relative contribution of the head and eyes, there were two types of blocks: Eyes-Visible and Eyes-Invisible, in which the looker wore a pair of sunglasses. In each block, the looker gazed at each object once in a random order (52 gaze trials per block). Each trial lasted 10 seconds. The looker kept staring at the target until a ring sound indicating the end of this trial. The looker then returned to resting state by watching the command computer's screen. Observers rotated their positions during the experiment, so that the combination of looker, eye-condition and position (center or peripheral) were balanced. Experiment 2 was identical to Experiment 1 except that only the Eyes-Visible condition was included. There were two Caucasian lookers (VO and HE) and 10 Caucasian and African-American observers, which were assigned into three sections. There were 12 blocks in each section.

In addition to the two formal experiments, we also invited 10 new lookers to perform the free looking task for 3 blocks (30 blocks in total), which were also recorded by the RGB-D camera. These were the training data of our computational models.

## 3     Appearance-based computational model

Here we construct a computational model that can infer a 3D gaze direction given a 2D image. In our model, the RGB-D data is used only for training purposes, while for prediction, only 2D images are given. In real world, we can look at a single point with infinite head-eye combinations. The final direction of gaze is determined by both the head orientation (pitch, yaw and roll) and the correction due to the eyes-gaze. In our setup, the final gaze direction is known, given the position of the trial target, and the head orientation is provided by the RGB-D sensor. Therefore, eyes-gaze direction can be deducted by subtracting head-orientation from the final gaze direction. Our Appear-Face-Eyes model, first associates the whole face appearance to the 3D head orientation, and then corrects for the final direction of gaze by associating the appearance of the eyes with the 3D eyes-gaze offset.

We compare the Appear-Face-Eyes model with two additional models: (1) The Appear-Face model associates the whole face appearance directly to the final gaze direction. In this model, information about the 3D head pose orientation and eyes-gaze appearance are disregarded. This model should capture the role of the whole face appearance in detecting the final direction of gaze; (2) The Kinect-Linear model is a linear regression model between the Kinect's head pose orientation and the final gaze direction. In this model all information about the face and eyes appearance are disregarded. This model provides a reference performance for the Appear-Face-Eyes model, which is trained on the Kinect's head pose orientation data.

The Appear-Face-Eyes model is based on a gaze detection method used in [22], in which grayscale image patches of faces were associated with the 2D direction vector of the gaze (the vector connecting the face center to the regarded target position in the 2D image). During training, a set of face image patches are observed and associated with the 2D direction of gaze corresponding to each of the faces. To detect the direction of gaze in a new query image, a small set of the most similar faces to the face in the query image, are extracted from the training set. The direction of gaze is then estimated as the weighted average of the gaze directions associated with the query face neighbors from the training set. This approach demonstrated full generalization across faces and directions of gaze.

Our model extends the above 2D model, to cope with direction of gaze in 3D while associating both head orientation and eyes-gaze direction with the 2D image of the face and eyes. The Kinect's RGB-D sensor combined with Microsoft's face tracking method provides us with the required 3D and 2D information we need for the supervised training phase. We utilize Microsoft Kinect's face tracker to extract both the 3D position and orientation of the head, as well as the position of several face features, whenever a human actor is looking at a given target. The 3D position of the target is also extracted from the Kinect's depth image. From both head and target positions we can calculate the 3D eyes-gaze correction as the difference between the head orientation and the final direction of regard from the face center to the target (Figure 2). The Kinect RGB-D sensor also provides a 2D color image, from which we extract image patches of the whole face and eyes region. In our model, the face appearance is associated with the 3D head orientation, and the eyes region appearance is associated with the 3D eyes-gaze correction to the target. We use Histograms of Gradients descriptors (HoG, [23]) for the representation of the face and eyes appearance, and quaternions for representing the 3D orientations. HoG descriptors are robust general-purpose image descriptors, which capture both edge and texture properties and allow spatial tolerance.

During detecting in a query gaze event, our model is applied to the 2D query image as follows: First, image patches around the face and eyes region are extracted, and HoG appearance descriptors are computed. The model then searches for the nearest neighbors from the training set for both face and eyes appearances. The 3D face orientation is estimated as the weighted average of the 3D head orientations associated with each of the face neighbors, where the weights are proportional to the neighbors' similarity with the query face. The 3D eyes-gaze correction is estimated similarly, using neighbors of the eyes region. Note that in our model similar eyes-gaze corrections, may be associated with different eyes region appearances which relate to different face orientations, as we do not rectify the eyes appearance back to frontal view from different head orientations. This may be a more realistic and cognitively plausible approach, similar to humans experiencing different face and eye appearances at different poses when observing other humans around them.

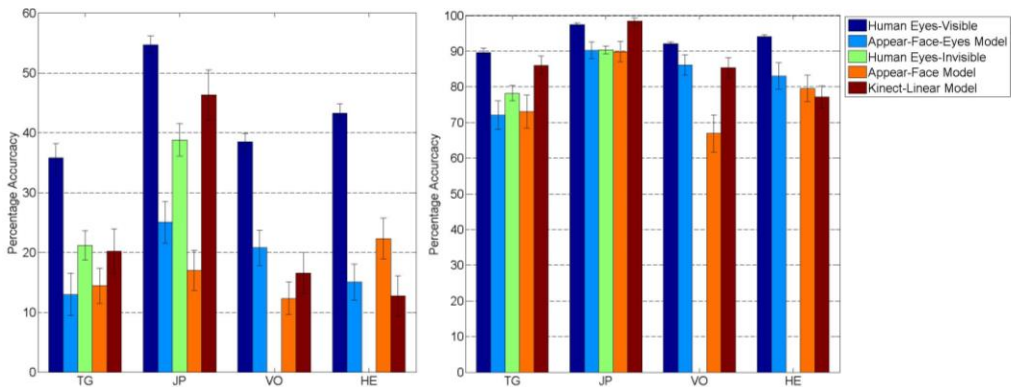

Figure 3: Accuracy graphs. Horizontal axis represents four different lookers. Different bars represent human (Eyes-Visible for all 4 lookers and Eyes-Invisible for TG and JP only) and models' results. (Left) The accuracy of getting the exact column and row right. (Right) The accuracy of allowing one row or one column off.

## 4 Results

The overall performance of humans and the models are shown in Figure 3 (left). The performance is measured as the percentage of trials in which humans or models get both the correct target row and column positions. Figure 3 (right) presents the percentage of trials in which the errors are no larger than one column or one row. This corresponds to about 12° of visual angle range around the true target. The results reflect several interesting patterns: (1) humans outperform all the models, suggesting that humans' gaze perception is indeed highly efficient. (2), there is a significant amount of across-looker variance. Both human and models are much better at reading JP's gaze than the rest of the lookers, showing that the models actually capture some intrinsic variances of human performance. (3) For both humans and models, most errors are within a 12° range. For models, this performance is very good compared with current reported accuracy on a similar task [24].

The contrast of Eyes-Visible and Eyes-Invisible conditions also reveals the importance of both the eyes and the gaze. The performance is much lower when the eyes are invisible (~15% drop in accuracy). Interesting, the performance difference between TG and JP do not change at all, suggesting the spurious performance of reading JP's gaze is primarily due to head-pose, instead of eyes. In addition, humans' Eye-Invisible performance is much similar to the models' performance. By using the Kinect head pose tracking, the performance reaches the human level. For the other more challenging appearance-based models, using head pose alone or using both head and eyes makes little difference, suggesting that the HoG descriptors we used here are indeed able to capture important visual information of the head pose but not the eyes.

Figure 3 plots the overall accuracy of each observer with the Eyes-Visible condition. It reveals a remarkable individual difference, varying from ~20% to ~65% (Note that there are 208 trials for each observers in Experiment 1, and 624 trials in Experiment 2. It is unlikely that this is just due to random variance). The individual difference of gaze perception has not been emphasized in previous studies. It is interesting to explore how this difference can impact subsequent social processes. For modeling purpose, this range of performance also indicates that the "optimal" performance of this task can be much higher than the averaged human performance.

We also explored the interaction between the looker's gaze direction and the observers' viewing positions. Figure 4 shows the accuracy of getting the row and the accuracy of getting the column right when the target is in the left most column and right most column across the different observer positions. The observers performs the best when the looker is gazing towards their direction, while their column accuracy drops significantly as the viewing direction of the looker is getting further towards a side view. This drop in performance does not occur for the row accuracy.

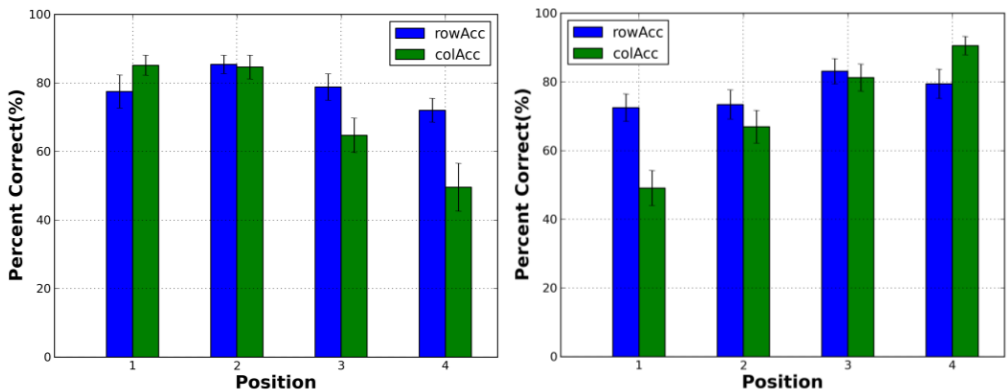

Figure 4: Position effects. (Left) The averaged accuracy of getting the row (blue) and column (green) right when the looker gazes at a target in the left most column. Observes sitting in Position 1, 2 on the left side of the table are much more accuracy at discriminating the columns, but not the rows. (Right) The averaged row and column accuracy when the looker gazes at a target in the right most column. Observers siting in Position 3 and 4 on the right side of the table are much more accurate at discriminating the column, but not the row.

Table 1: Bias and standard deviation. colBias represents the bias in the column (left-right) direction. colStd represents the standard deviation in the column direction. rowBias represents the bias in the row (up-down) direction. rowStd represents the standard deviation in the row dimension. It is very clear that the models' standard deviation is comparable, even smaller than human. However, the bias of the models are much larger and inconsistent.

| Type | Looker | colBias | colStd | rowBias | rowStd |
|---|---|---|---|---|---|
| Model Appear-Face | HE | 0.14 | 5.32 | -2.73 | 4.93 |
| | JP | -2.74 | 3.89 | 0.69 | 3.76 |
| | TG | 4.92 | 4.38 | -3.61 | 4.91 |
| | VO | -3.63 | 4.81 | 6.96 | 3.87 |
| Model Appear-Face-Eyes | HE | 0.77 | 4.90 | -1.50 | 4.32 |
| | JP | -1.35 | 4.39 | -2.66 | 4.19 |
| | TG | 5.96 | 4.36 | -5.48 | 4.12 |
| | VO | -2.65 | 4.73 | 3.42 | 3.85 |
| Model Kinect-Linear | HE | 5.75 | 4.32 | -4.66 | 3.53 |
| | JP | 0.53 | 3.11 | 0.79 | 3.52 |
| | TG | 1.62 | 5.69 | -0.24 | 4.24 |
| | VO | -3.45 | 4.68 | 3.31 | 3.35 |
| Human Eyes-Visible | TG | 3.18 | 6.09 | -2.82 | 4.53 |
| | JP | 0.67 | 5.51 | 0.53 | 3.47 |
| | VO | -0.57 | 8.88 | -0.73 | 4.83 |
| | HE | 0.76 | 8.18 | -1.76 | 4.59 |
| Human Eyes-Invisible | TG | 3.76 | 7.42 | -1.81 | 7.12 |
| | JP | 2.00 | 6.79 | 0.25 | 5.10 |

**Bias and Variance**

Here we further explored the results by analyzing the distributions of humans and models responses to a given target. Two summary statistics are chosen: Bias and standard deviation (SD). For both humans and models, a poor performance could be due to a large systematic bias, or a large variance.

Bias measures the mean of the response relative to the true position. For the column dimension (left-right), a positive bias is defined as the mean is biased toward the peripheral of the looker's visual field (e.g., for columns on the left side of the looker, a positive bias indicates the perceived gaze direction moves toward the left even further. For the right side, a positive bias is more toward the right). For the row dimension (up-down), a positive bias means that the perceived gaze direction moves downward (further away from the resting direction).

Table 1 depicts the bias and variance along the left-right and up-down dimension. What really strikes us is that the models' SDs are almost as good as those of humans. The relatively small SD indicates that given the same actor, when he/she looks at the same object in different blocks, the models' predictions are consistent. What makes the models' perform worse is the lack of small bias across lookers and dimensions (row and column). Both bias and variances had been discussed in previous studies (e.g. [9], [10]). However, since then, much attention has been given to the small variance of gaze perception. Our contrast of human perception and computer vision suggests that at least in the context of free looking, it is the small bias that is most difficult to achieve.

# 5  Discussion

In this paper, we explored the acuity of reading others' gaze in a "free looking" task with both human participants and computer vision models. The guide lines for this study are to keep the ecological validity as high as possible, while maintaining quantitative precision. The ecological validity is emphasized, as the results should serve for understanding social interactions in real scenes.

One straightforward result is that the human accuracy of discriminating targets 8°-10° of visual angle apart is around 40% in a free looking gaze task. However, the most challenging part of this project is not to obtain this accuracy, but how to interpret it. By comparing with several computer vision models, it turns out that human visual perception achieves a remarkable level of performance, considering the computational challenge involved. More interestingly, by comparing model performance with the Eyes-Visible, and Eyes-Invisible conditions, we found that while a general-purpose appearance descriptor captures useful information about the head pose, it's processing of the eyes region makes very little additional contributions. This result indicate that in a free-looking paradigm, in which the appearance of the eyes region varies dramatically for the observer, a better representation of the eyes may be required. This representation should still capture the inter-pose variations, since assuming that eyes appearance should always be represented and analyzed as rectified to a frontal head pose is less likely to be cognitively plausible.

In addition, we also observed a significant across-looker variance. Both the humans and models are much better at reading one looker (JP)'s gaze. Since our models are only good at detecting head poses, it suggests that this across-looker variance is primarily due to the head pose associated with each target. This is consistent with the behavioral data showing that this difference was still present when the looker's eyes were invisible in Experiment 1. Taken together, the current research shows how the computational process of human social perception can be analyzed in depth by combining cognitive paradigms with computer vision.

## Acknowledgments

This material is based upon work supported by the Center for Brains, Minds and Machines (CBMM), funded by NSF STC award CCF-1231216, and NSF National Robotics Intuitive.

## References

[1]  T. Gao, G. E. Newman, and B. J. Scholl, "The psychophysics of chasing: A case study in the perception of animacy.," *Cogn. Psychol.*, vol. 59, no. 2, pp. 154–79, Sep. 2009.
[2]  C. L. Baker, R. Saxe, and J. B. Tenenbaum, "Action understanding as inverse planning.," *Cognition*, vol. 113, no. 3, pp. 329–49, Dec. 2009.
[3]  M. Tomasello, *The cultural origins of human cognition*. Harvard University Press, 1999.
[4]  M. Scaife and J. S. Bruner, "The capacity for joint visual attention in the infant," *Nature*, vol. 253, pp. 265–266, 1975.
[5]  B. D'Entremont, S. M. J. Hains, and D. W. Muir, "A demonstration of gaze following in 3- to 6-month-olds," *Infant Behav Dev*, vol. 20, no. 4, pp. 569–572, 1997.
[6]  D. Premack and G. Woodruff, "Does the chimpanzee have a theory of mind?," *Behav. Brain Sci.*, 1978.
[7]  R. Saxe, S. Carey, and N. Kanwisher, "Understanding other minds: linking developmental psychology and functional neuroimaging.," *Annu. Rev. Psychol.*, vol. 55, pp. 87–124, Jan. 2004.
[8]  R. Flom, K. Lee, and D. Muir, *Gaze-following: It's development and significance*. Mahwah: Lawrence Erlbaum Associates, 2007.
[9]  J. Gibson and A. Pick, "Perception of another person's looking behavior," *Am. J. Psychol.*, 1963.
[10] M. Cline, "The perception of where a person is looking," *Am. J. Psychol.*, 1967.
[11] L. A. Symons, K. Lee, C. C. Cedrone, and M. Nishimura, "What are you looking at? Acuity for triadic eye gaze," *J. Gen. ...*, vol. 131, no. 4, 2004.
[12] S. R. Schweinberger, N. Kloth, and R. Jenkins, "Are you looking at me? Neural correlates of gaze adaptation.," *Neuroreport*, vol. 18, no. 7, pp. 693–6, May 2007.
[13] D. Todorović, "Geometrical basis of perception of gaze direction.," *Vision Res.*, vol. 46, no. 21, pp. 3549–62, Oct. 2006.
[14] B. Stiel, C. Clifford, and I. Mareschal, "Adaptation to vergent and averted eye gaze," *J. Vis.*, 2014.

[15] S. W. Bock, P. Dicke, and P. Thier, "How precise is gaze following in humans?," *Vision Res.*, vol. 48, no. 7, pp. 946–57, Mar. 2008.
[16] R. H. S. Carpenter, *Movements of the Eyes*. London: Pion Limited, 1988, p. 593.
[17] E. Murphy-Chutorian and M. M. Trivedi, "Head pose estimation in computer vision: A survey," *Pattern Anal. Mach. Intell. IEEE Trans.*, vol. 31, no. 4, pp. 607–626, 2009.
[18] D. W. Hansen and Q. Ji, "In the eye of the beholder: a survey of models for eyes and gaze.," *IEEE Trans. Pattern Anal. Mach. Intell.*, vol. 32, no. 3, pp. 478–500, Mar. 2010.
[19] A. Gee and R. Cipolla, "Determining the gaze of faces in images," *Image Vis. Comput.*, vol. 12, no. 10, pp. 639–647, Dec. 1994.
[20] J. Odobez and K. F. Mora, "Person Independent 3D Gaze Estimation From Remote RGB-D Cameras," *Int. Conf. Image Process.*, 2013.
[21] S. Russell, P. Norvig, J. Canny, J. Malik, and D. Edwards, *Artificial intelligence: a modern approach*. 1995.
[22] S. Ullman, D. Harari, and N. Dorfman, "From simple innate biases to complex visual concepts," *Proc. Natl. Acad. Sci.*, vol. 109, no. 44, pp. 18215–18220, Sep. 2012.
[23] N. Dalal and B. Triggs, "Histograms of Oriented Gradients for Human Detection," in *Proceedings of Computer Vision and Pattern Recognition*, 2005, pp. 886–893.
[24] J. Odobez and K. F. Mora, "Person Independent 3D Gaze Estimation From Remote RGB-D Cameras," *Int. Conf. Image Process.*, 2013.